\pdfoutput=1
\documentclass{article}
\usepackage{soul}
\usepackage{lineno,hyperref}
\usepackage{multirow}
\usepackage[final,nonatbib]{nips_2016}
\usepackage[numbers]{natbib}

\usepackage[utf8]{inputenc} 
\usepackage[T1]{fontenc}    
\usepackage{hyperref}       
\usepackage{url}            
\usepackage{booktabs}       
\usepackage{amsfonts}       
\usepackage{nicefrac}       
\usepackage{microtype}      
\usepackage{amsmath}
\usepackage[pdftex]{graphicx}

\title{Detection of concealed cars in complex cargo X-ray imagery using deep learning}

\author{
  Nicolas~Jaccard$^{1}$ \quad Thomas W.~Rogers$^{1,3}$ \quad Edward J.~Morton$^{2}$  \quad Lewis D.~Griffin$^{1*}$\\\\
  $^1$Department of Computer Science, University College London, London, UK\\
  $^2$Rapiscan Systems Ltd., Stroke-on-Trent, UK\\
  $^3$Department of Security and Crime Sciences, University College London, London, UK\\\\
  \small{$^{*}$Corresponding-author: l.griffin@cs.ucl.ac.uk}
}

\begin{document}

\maketitle

\begin{abstract}
Non-intrusive inspection systems based on X-ray radiography techniques are routinely used at transport hubs to ensure the conformity of cargo content with the supplied shipping manifest. As trade volumes increase and regulations become more stringent, manual inspection by trained operators is less and less viable due to low throughput. Machine vision techniques can assist operators in their task by automating parts of the inspection workflow. Since cars are routinely involved in trafficking, export fraud, and tax evasion schemes, they represent an attractive target for automated detection and flagging for subsequent inspection by operators. In this contribution, we describe a method for the detection of cars in X-ray cargo images based on trained-from-scratch Convolutional Neural Networks. By introducing an oversampling scheme that suitably addresses the low number of \emph{car} images available for training, we achieved 100\% \emph{car} image classification rate for a false positive rate of 1-in-454. Cars that were partially or completely obscured by other goods, a modus operandi frequently adopted by criminals, were correctly detected. We believe that this level of performance suggests that the method is suitable for deployment in the field. It is expected that the generic object detection workflow described can be extended to other object classes given the availability of suitable training data.
\end{abstract}

\section{Introduction}
Non-Intrusive Inspection (NII) systems are routinely used at transport hubs to scan the content of cargo containers, and ensure their compliance with both the shipping manifest and transport regulations, without disrupting the flow of commerce~\cite{Sarathy2006}. NII systems use radiation such as fast neutrons, gamma-rays or most commonly X-rays to image containers~\cite{Cutmore2010,McDaniel2005a,liu2008comparison}. Currently, X-ray transmission images are inspected by human operators who search for anomalies or discrepancies with the shipping manifest~\cite{Tuszynski2013b}. 

Despite ambitious plans to scan all cargo entering the United States~\cite{archick2010us}, it is not feasible to image every container due to the ever-increasing international trade volumes~\cite{Asariotis2013}, let alone visually inspect all images hypothetically produced in the process. Container targeting is routinely carried out by risk analysis based on information such as origin, destination, and declared content~\cite{King2005,Weele2010}. This approach limits the number of containers to image to those deemed ``high risk'' and thus greatly reduces the impact on the flow of commerce. However, the number of images to manually inspect remains overwhelmingly high, a trend compounded by the recent deployment of high throughput X-ray scanners capable of imaging cargo transported by rail at speed.

The application of machine vision methods to X-ray cargo images present many advantages over manual inspection, including high throughput through automation, consistency, scalability, and resistance to corrupt operation. However, as discussed in related work (Sec.~\ref{sec:relatedWork}), there is little published work on automated cargo X-ray image processing, potentially due to the difficulty of obtaining suitably large labeled datasets~\cite{Mery2014}. 

This contribution describes highly accurate algorithms for the detection of cars in complex X-ray cargo imagery. The automatic and reliable detection of cars is highly desirable because they are routinely involved in export fraud, tax evasion schemes, and trafficking activities~\cite{Aronowitz1996,clarke2003international,Clarke2010}. Two main challenges are addressed: i) detection of cars that were intentionally obscured by other goods in order to minimise risk of detection by imaging and physical inspection; and ii) minimise the false alarm rate on \emph{non-car} images that frequently contain ``car-like'' patterns. The latter point is particularly important for deployment in the field as unjustified false alarms could lead to operators ignoring the output of the automated detection scheme, or discourage its use altogether.

Results for preliminary \emph{car} image classification experiments were previously presented at a conference~\cite{Jaccard2014b}. However, the dataset used for evaluation was small and did not contain challenging adversarial examples. Moreover, the classification scheme was based around fixed image features only (intensity values, BIFs, oBIFs). This new contribution explores, for the first time, the classification of large X-ray cargo images using CNNs (either trained from scratch on X-ray cargo imagery or pre-trained on natural images) and compare their performance with that of the aforementioned fixed features. The dataset used for evaluation is also  much larger and contains adversarial examples where cars are partially or completely obscured by other goods. Furthermore, additional experiments were carried out to better characterise the resilience of the CNN classification scheme to obscuration. 

The structure of this paper is as follows. First, the formation process and properties of X-ray transmission images are briefly introduced in section~\ref{sec:xrayImages}. In section~\ref{sec:relatedWork}, related work from the literature is discussed. The methods underpinning the experiments carried out in this study are then outlined in section~\ref{sec:method} while results of performance evaluation experiments comparing Convolutional Neural Networks (CNNs) with other types of features for the detection of cars in stream-of-commerce (SoC) images are presented in section~\ref{sec:results}. Finally, findings, limitations, and avenues of future research are discussed in section~\ref{sec:conclusion}.

\section{X-ray transmission image formation and properties}
\label{sec:xrayImages}
\begin{figure}[h]
\centering
\includegraphics[width=0.98\linewidth]{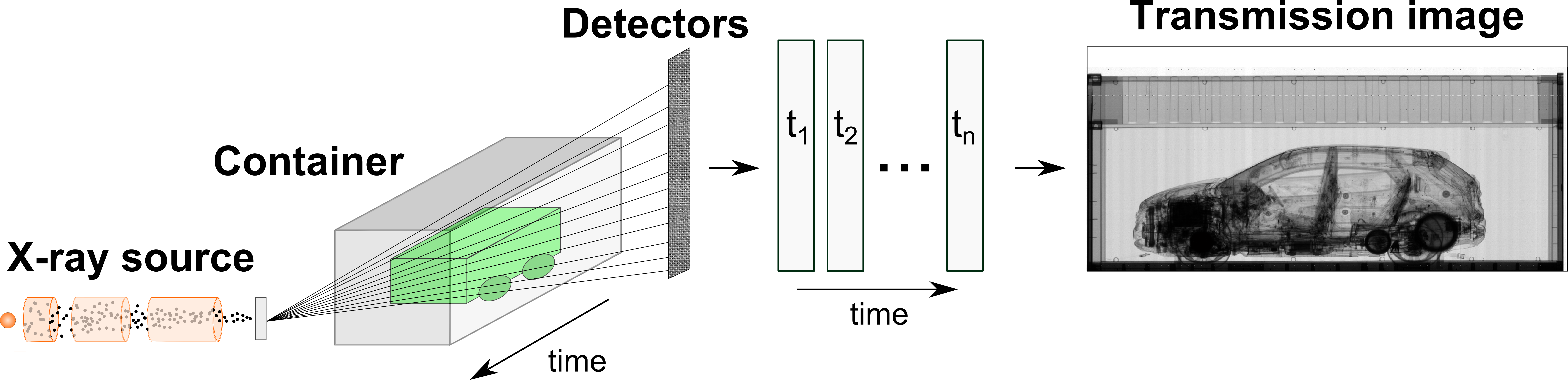}
\caption{Illustration of the X-ray image formation and acquisition processes. Photons emitted by an X-ray source interact with a container and its content, leading to a signal attenuation measured by detectors placed behind the container. By moving the container or the detector, attenuations are determined spatially and are be mapped to pixel values to produce an X-ray transmission image}
\label{fig:XRayPrinciples}
\end{figure}

The two main components of a typical X-ray cargo scanner are an X-ray source and an array of detectors located behind the scanned object (Fig.~\ref{fig:XRayPrinciples}). The fan-shaped X-ray beam (large vertical and small horizontal spread) is matched by the tall, narrow geometry of the detector array. The fraction of photons absorbed or scattered by the container and its content, is determined by measuring the number of photons incident on the detectors. The signal attenuations measured at different locations on the detector array are then mapped to pixel values, forming an X-ray transmission image where low attenuation regions (e.g. air) and regions containing dense objects have high and low pixel values, respectively. Due to the narrow geometry of the detector array, this acquisition process has to be repeated multiples times by moving either the container (portal configuration) or the source and detector array (gantry configuration). Individual column images are then assembled to form the final image.

X-ray transmission images differ significantly from visible spectrum photographs. In general, X-ray images have a skewed perspective due to the position of the source relative to the detector and scanned object, contain partially overlapping translucent objects, are cluttered, and are highly noisy~\cite{Zhang2014,Ba2011,McDaniel2005a}. 

\section{Related work}
\label{sec:relatedWork}
Methods have been described to enhance images for improved target detection by operators, and to fully replace operators with detection algorithms. For enhancements, Retinex filters, false colour mapping, and the fusion of images acquired at two different energies (dual-energy X-ray imaging, e.g. for material discrimination based on atomic number) have been explored~\cite{Abidi2005,Woodell2004c,Abidi2006,Ogorodnikov2002a}. For automated detection, bag of words (BoW) representations classified using support vector machines (SVM) were used for the analysis of 2D and 3D X-ray scans of baggage containing objects of interests such as firearms and mobile phones~\cite{Turcsany2013,Ba2011,Flitton2015}. These studies report impressive performance, which is in part made possible by the relatively constrained process of baggage scanning: scene dimensions and complexity are both bounded by the small dimensions of a bag. Multi-view (potentially volumetric), multi-energy, and high resolution imaging enable discriminating between threats and legitimate objects, with the latter being mostly identical across different baggage.

In contrast, the detection of threats and anomalies in X-ray cargo imagery is significantly more challenging. Scenes tend to be very large and complex with little constraints on the arrangement and packing of goods. Scanning is usually limited to a single view and the spatial resolution is much lower than in baggage, making it especially difficult to resolve and locate small anomalous objects. Moreover, a very high fraction of items packed in baggage are well-cataloged (e.g. clothing), whereas potentially anything can be transported in a container making it impractical to learn the appearance of frequent legitimate objects to facilitate the detection of threats. For these reasons, the performance reported for cargo imagery is usually low.

Zhang \emph{et al.}~\cite{Zhang2014} built a so-called ``joint shape and texture model'' of X-ray cargo images based on BoW extracted in superpixel regions. Using this model, images were classified into 22 categories depending on their content (e.g. car parts, paper, plywood). The results highlighted the challenges associated with X-ray cargo image classification, with only 51\% of images being assigned to the correct category. In another effort to develop an automated method for the verification of cargo content in X-ray images, Tuszynski \emph{et al.}~\cite{Tuszynski2013b} developed models based on the log-intensity histograms of images categorized into 92 high-level HS-codes (Harmonized Commodity Description Coding System). A city block distance was used to determine how much a new image deviates from training examples for the declared HS-code. Using this approach, 31\% of images were associated with the correct category, while in 65\% of cases the correct category was amongst the five closest matching models.

With around 20\% of cargo containers being shipped empty, it would be of interest to automatically classify images as empty or non-empty in order to facilitate further processing (e.g. avoid processing empty images with object-specific detectors) and to prevent fraud. Rogers \emph{et al.}~\cite{rogers2015detection} described a scheme where small non-overlapping windows were classified by a Random Forest (RF) based on multi-scale oriented Basic Image Features (oBIFs) and intensity moments. In addition, window coordinates were used as features so that the classifier would implicitly learn location-specific appearances. The authors reported that 99.3\% of SoC non-empty containers were detected as such for a 0.7\% false alarm rate and that 90\% of synthetic images (where a single object equivalent to 1L of water was placed) were correctly classified as empty for 0.51\% false alarms. The same problem was tackled by Andrews and colleagues~\cite{andrews2016detecting} using an anomaly detection approach; instead of implementing the empty container verification as a binary classification problem, a ``normal'' class is defined (either empty or non-empty containers) and new images are scored based on their distance from this ``normal'' class. Features of markedly down-sampled images ($32\times 9$ pixel) were extracted from the hidden layers of an auto-encoder and classified by a one-class SVM, achieving 99.2\% accuracy when empty containers were chosen as the ``normal'' class and non-empty instances were considered as anomalies.

Representation-learning is an alternative to classification based on designed features, whereby the image features that optimise classification are learned during training. CNNs, often referred to as deep learning, are representation-learning methods~\cite{LeCun2015} that were recently shown to significantly outperform other machine vision techniques in many applications, including large-scale natural image classification~\cite{He2015}. While most examples of applications to X-ray imagery to date have been limited to medical data~\cite{CERNAZANU-GLAVAN2013}, Ak{\c{c}}ay~\emph{et al.}~\cite{akccaytransfer} recently demonstrated the use of CNNs for baggage X-ray image classification. As there was insufficient training data to train a network from scratch, the authors fine-tuned a variant of the AlexNet architecture~\cite{krizhevsky2012imagenet} that was pre-trained on ImageNet, a dataset of natural images. This approach significantly outperformed prior work in the field, indicating that features learned from natural images do indeed transfer, at least to a certain degree, to X-ray images. 

To our knowledge, CNNs have not been applied to X-ray cargo imagery. In this contribution, we compare CNNs with other types of features and determine whether trained-from-scratch models (e.g. trained only on X-ray images) perform better than pre-trained networks.

\section{Method}
\label{sec:method}
\subsection{Dataset}
X-ray transmission images of SoC cargo containers (typically 20 or 40 foot long) and tankers transported on railway carriages were acquired using a Rapiscan Eagle\textregistered R60 rail scanner equipped with a 6 MV linac source. Image dimensions vary between $1290\times 850$ and $2570\times 850$ pixel depending on the type of cargo and container size, with a pixel size of $\approx 6$ mm pixel$^{-1}$ in the horizontal direction. The raw images are greyscale with 16-bit precision.

For the purpose of this work, images containing \emph{at least} one car (\emph{car} images) are taken as the positive class and images not containing any car (\emph{non-car} images) as the negative class. The dataset contains 79 \emph{car} images for a total of 192 individual cars. \textit{Car} images can be broadly divided into 5 categories: (i) a single car on its own in a small container (20 ft long), (ii) two cars in a large container (40 ft long), (iii) multiple cars stacked in a container, including one at an angle, (iv) a single car next to unrelated goods (no overlap), (v) one or two cars placed in-front or behind other goods (partial or complete occlusion). The specific car models and manufacturers were unknown, however based on visual appearances sedans, SUVs, compacts, and sports cars were present in the dataset. 

\textit{Non-car} images were randomly sampled from SoC images acquired over the course of several months. These images can be of cargo containers and tankers, with the first type being the most frequent. The nature of the cargo loads varies greatly from a container to another and include pallets of commercial goods, industrial equipment, household items, and bulk materials. Approximately 20\% of the containers imaged were empty. \textit{Non-car} images also include other types of vehicles such as vans, motorbikes, and industrial vehicles (e.g. tractors, bulldozers).

\subsection{Image pre-processing}

Prior to classification, X-ray transmission images were pre-processed as previously described by Rogers \textit{et al.}~\cite{rogers2014reduction,rogers2015detection}. Black stripes resulting from source misfires or faulty detectors were first removed. Variations in the source intensity and sensor responses were corrected by column-wise pixel intensity normalisation based on air attenuation values, which are considered invariant. Erroneous isolated pixels (e.g. excessively bright or dark) were replaced by the median of their neighbourhood. For certain experiments, the log transform of images was also computed as it is frequently used to facilitate the detection of concealed items by operators and was also previously employed for the automated classification of cargo images by Tuszynski and colleagues~\cite{Tuszynski2013b}.

\subsection{Classification scheme}
\begin{figure}
\centering
\includegraphics[width=0.95\linewidth]{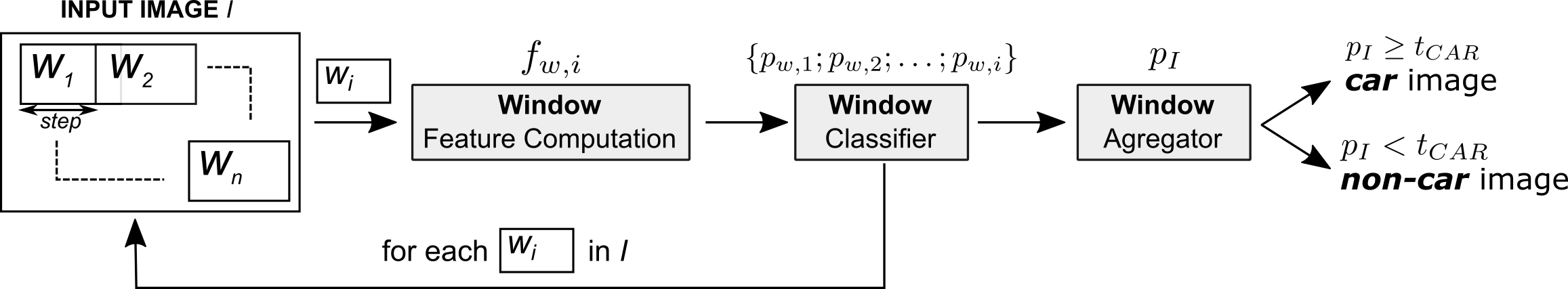}
\caption{A window-based scheme for the classification of large X-ray cargo images. Windows are densely sampled from large input images and their features computed, based on which their ``car-likeness'' score is assigned by a window classifier. An image score is computed as the maximum window score across all windows of an image. Image class label (\emph{car} or \emph{non-car}) is obtained by thresholding of the image score.}
\label{fig:Scheme_General}
\end{figure}
The detection of cars in X-ray images was implemented as a binary classification task (Fig.~\ref{fig:Scheme_General}). A window-based approach was taken enabling i) to process optimally small sub-images for high classification performance as well as low computational time and memory consumption, and ii) to obtain approximate localisation of car-containing regions. Each window $w_{i}$, densely sampled from an image $I$, was classified and associated with a ``car-likeness'' score $p_{w,i}$. The image score $p_{I}$, which is indicative of the confidence that the image contains \emph{at least} one car, was given by the maximum value of $p_{w,i}$ across all $w_{i}$ of $I$. The image was classified as \textit{car} if $p_{I}{\geq}t_{CAR}$, and \textit{non-car} otherwise. $t_{CAR}$ is a tunable threshold parameter that defines the balance between detection and false alarm rates.

Two types of windows were evaluated: square $512{\times}512$ pixel and rectangular {$350{\times}1050$ pixel. The latter corresponded to the average size of cars in the training set and can be interpreted as a geometric prior. In all cases, windows were sampled with a stride of 32 pixels and 64 pixels for training and inference, respectively. 

Heatmaps for classification visualisation were generated by mapping the mean window response at all image locations to pixel values. Such visualisations are essential to clarify the decision of the automated detection scheme and to enable verification by the operator before deciding whether further actions (e.g. physical inspection) are required.

Windows were classified by RF, SVM or logistic regression (for CNNs only) based on pixel intensity, fixed geometric image descriptors (BIFs), learned visual words (Pyramid Histograms Of Visual Words, PHOW), and features extracted from CNNs.
\subsection{Window classification using Random Forest and Support Vector Machines}
\label{sec:classifiers}
%
For this work, an open-source implementation of Random Forest for MATLAB was employed\footnote{https://code.google.com/p/randomforest-matlab/ - Last accessed 31.05.2016}. If not otherwise stated, classification was carried out using 40 trees, randomly sampling the square root of the total number of features at each split during tree building, and using equal weights for the two classes. For each window, the classifier outputs the ``car-likeness'' score $p_{w,i}$ computed as the fraction of trees voting for the \textit{car} class.

Classification using linear SVMs was implemented using MATLAB's built-in functions. The box-constraint (or regularisation) parameters $C$ and the kernel scale $\gamma$ were tuned empirically. The ``car-likeness'' score $p_{w,i}$ was computed using a function that maps uncalibrated SVM scores to posterior probabilities. As proposed by Platt~\cite{Platt}, a sigmoid was used as mapping function and parameters were estimated post-training using 10-fold cross validation.

In addition to RF and SVM, softmax was also used for classification using CNNs as described in section~\ref{sec:CNNs}.
\subsection{Feature computation}
The simplest type of features assessed for \emph{car} image classification was intensity values (Sec.~\ref{sec:intensityFeatures}). More advanced descriptors included oBIFs (fixed geometric features, sec.~\ref{sec:BIFs}) and PHOW (learned visual words, sec.~\ref{sec:PHOW}). CNNs for feature computation and classification are described in section~\ref{sec:CNNs}.

\subsubsection{Intensity features}
\label{sec:intensityFeatures}
Intensity features were encoded in multi-scale 256-bin histograms. Input images were blurred by convolution with a Gaussian kernel of standard deviation equal to 1, 2, 4, and 8. The resulting feature vector was 1024-dimensional. Histograms of intensity features were computed efficiently for a large number of windows using the integral histogram method described by Porikli~\cite{Porikli2005}.

\subsubsection{oriented Basic Image Features}
\label{sec:BIFs}
BIFs encode textural information by classifying pixels of an image into one of seven categories according to local symmetry~\cite{griffin2009basic}. BIFs were computed based on the response to a bank of derivative-of-Gaussian (DtG) filters~\cite{griffin2009basic,griffin2012improved}. The scale-normalized response $s_{ij}$ to the $ij$-th DtG $G_{ij}$ of scale $\sigma_{B}$ is shown in equation~ \ref{eq:normalisedResponse}.

\begin{equation}
\label{eq:normalisedResponse}
s_{ij}= \sigma_{B}^{i+j}G_{ij}\ast I
\end{equation}
Intermediate terms are then calculated pixel-wise: $\lambda$ (equation~\ref{eq:BIF_lambda}) is the scale-normalised image Laplacian and $\gamma$ (equation~\ref{eq:BIF_gamma}) is a measure of the variance over directions of the second directional derivative.
\begin{equation}
\label{eq:BIF_lambda}
\lambda = s_{20}+s_{02}
\end{equation}
\begin{equation}
\label{eq:BIF_gamma}
\gamma = \sqrt{(s_{20}+s_{02})^2+4s^{2}_{11}}
\end{equation}
The BIF value for a pixel is an integer between 1 and 7 given by the index of the largest of the following quantities: $\{\epsilon s_{00},\sqrt{s^{2}_{10}+s^{2}_{01}},\lambda,-\lambda,\frac{\gamma+\lambda}{\sqrt{2}},\frac{\gamma-\lambda}{\sqrt{2}},\gamma\}$, with $\epsilon$ being a threshold parameter that dictates when a pixel is considered `flat' (i.e. with no strong local structure), which is one type of BIF. The remaining six BIFs are slopes, dark blobs, bright blobs, dark lines, bright lines, and saddle-like (Fig.~\ref{fig:Scheme_BIFs}).

\begin{figure}[h]
\centering
\includegraphics[width=0.95\linewidth]{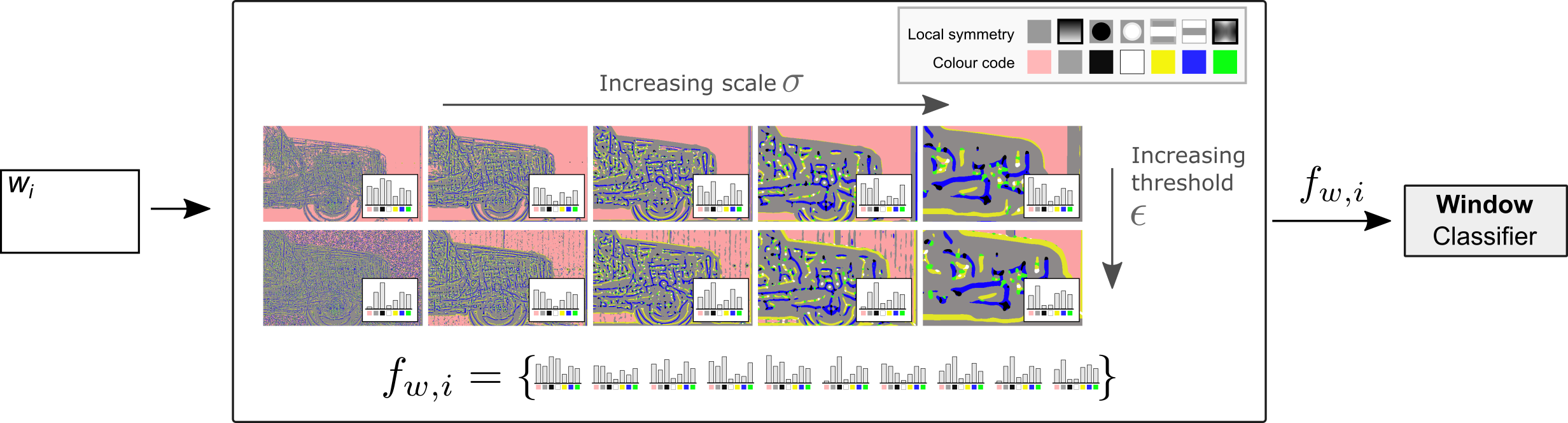}
\caption{Computation of oriented Basic Image Features for window classification. oBIFs for the input window are computed at multiple scales and for different threshold values. Histograms for each combination of parameters are constructed and concatenated to produce the window feature vector. For clarity, orientation quantization is omitted from the schematic.}
\label{fig:Scheme_BIFs}
\end{figure}

The BIF formulation can be extended by additionally determining the quantized orientation of rotationally asymmetric features~\cite{Newell2011}. This extended formulation, termed oriented Basic Image Features (oBIFs), has 23 features in total; with dark lines, light lines, and saddle-like types having 4 unpolarised orientations, while the slope type has 8 polarised directions. Implementations of both BIFs and oBIFs in MATLAB and Mathematica are available online~\cite{Griffin2015}.

oBIFs were computed at four scales ($\sigma_{B}{=}\{0.7, 1.4, 2.8, 5.6\}$) for two threshold parameters ($\gamma{=}\{0.011, 0.1\}$). oBIFs were encoded in histograms of 23 bins per scale and per threshold value, resulting in 184-dimensional feature vectors per window. As for intensity features, oBIFs histogram construction for multiple windows was carried out efficiently using the integral histogram method~\cite{Porikli2005}.
\subsubsection{Pyramid Histograms Of visual Words}
\label{sec:PHOW}
PHOW are a multi-scale extension of dense SIFT (Scale-Invariant Feature Transform) proposed by Bosch et \textit{al.} \cite{bosch2006scene,bosch2007image}. Whereas sparse SIFT approaches compute scale and rotation-invariant image descriptors based on local gradients at keypoint locations~\cite{lowe2004distinctive}, dense SIFT features are computed for each pixel or on a regular grid with constant spacing~\cite{dalal2005histograms}. The latter approach makes SIFT descriptors suitable for classification tasks where keypoints are not reliably detected or not consistent between the images considered, which is the case for X-ray cargo images.

PHOW computation (Fig.~\ref{fig:Scheme_PHOW}) consists of three steps : i) dense SIFT computation, ii) visual words quantization, and iii) spatial visual word histogram computation. SIFT descriptors were extracted at each location of a regular grid with a step of 3 pixels. SIFT descriptors are spatial histograms of image gradient with 8 orientation bins and arranged in $4{\times}4$ spatial bins centred at each grid location, producing a 128-dimension feature vector per location. This extraction step was carried out at four different scales (4, 6, 8, and 10 pixels) by varying the dimensions of the spatial bins. Images were smoothed prior to computation, with Gaussian kernels of standard deviation equal to the scale divided by 6. Descriptors were then quantized into 300 visual words that were learned by k-means clustering of training image descriptors. A two level pyramid histogram of visual words ($2{\times}2$ and $4{\times}4$ spatial bins) was constructed across all grid locations and scales, resulting in 6000-dimensional feature vectors for each window.
%

\begin{figure}[h]
\centering
\includegraphics[width=0.95\linewidth]{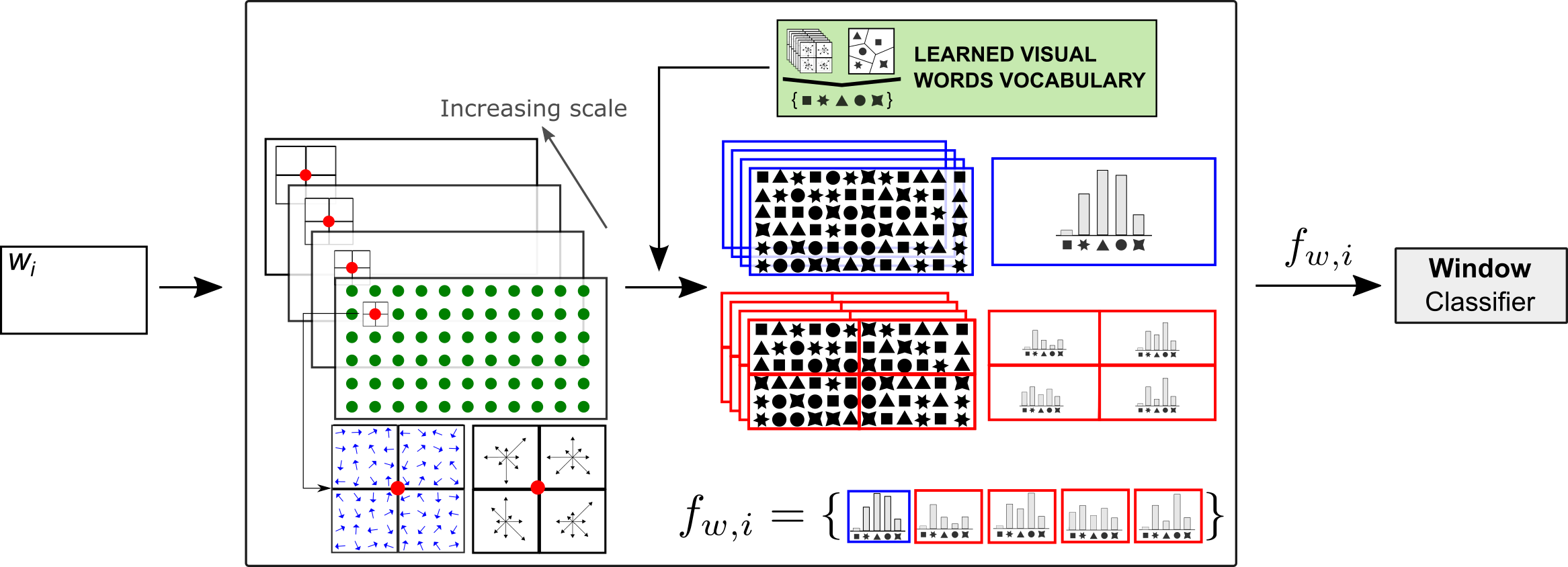}
\caption{Computation of PHOW features for window classification. SIFT descriptors are extracted at multiple scales before being quantized into visual words. A two level pyramid histogram of visual words is the constructed across scales. The feature vector is obtained by concatenation of all individual visual word histograms.}
\label{fig:Scheme_PHOW}
\end{figure}
\subsection{Convolutional Neural Networks}
\label{sec:CNNs}
CNNs were implemented using the MatConvNet library~\cite{vedaldi2014matconvnet}. Two types of network were evaluated, both based on the very deep architectures proposed by Simonyan and Zisserman~\cite{Simonyan2014}. The first one is a 11-layer architecture (8 convolutional layers and 3 full-connected layers), while the second is a 18-layer architecture (16 convolutional layers and 3 fully-connected layers). In both cases, all filters in the convolutional layers had $3{\times}3$ dimensions. Details of the architectures can be found in supplementary materials. The networks were regularised by batch normalisation, whereby the mean and variance of layer inputs are fixed~\cite{Ioffe2015}. Batch normalisation performed significantly better than the conventional regularisation approach that uses dropout layers~\cite{Srivastava2014}.

At the start of training, the learning rate was set to $10^{-4}$ and then to $10^{-5}$ when the validation error stopped decreasing. Weight decay was fixed at $5{\times}10^{-4}$. The average image computed over the training set was subtracted from all input images. When window classification was carried out solely based on CNNs, the ``car-likeness'' score $p_{w,i}$ was given directly by the output of the softmax classifier. In some experiments, features extracted from the first or second connected layers (FC1 and FC2, respectively) were classified using Random Forest or SVM classifiers as outlined in \ref{sec:classifiers}. Only $512{\times}512$ square windows were considered for classification using features extracted from CNNs. In order to make the memory footprint suitable for GPU processing, input images were first down-sampled to $256{\times}256$ pixels and converted to 8-bit precision.

In addition to models trained from scratch on windows sampled from X-ray cargo imagery, transfer learning was also evaluated. Window features extracted from the FC1 and FC2 layers of the VGG-VD-19 model~\cite{Simonyan2014} pre-trained on ImageNet were classified using Random Forest and SVM classifiers. As VGG-VD-19 expects $224{\times}224$ pixel RGB images as input, the grayscale channel of input X-ray images was replicated twice and downsampled, resulting in 3-channel $224{\times}224$ pixel images.

\subsection{Car oversampling}
While potentially millions of \emph{non-car} windows examples can be sampled from the SoC dataset, there are only a total of 192 individual cars. Training a balanced classifier (i.e. 192 windows for each classes) would certainly lead to poor performance and generalisation. A similar outcome would be expected if a classifier was trained on a severely imbalanced dataset containing significantly more \emph{non-car} examples. Such issues are frequently encountered in machine learning and more recently with CNNs where performance and generalisation is contingent on the availability of suitably large training datasets. Dataset augmentation by sampling random crops of input images at training was shown to significantly reduce CNN overfitting in large scale image classification tasks~\cite{krizhevsky2012imagenet}. A similar approach was taken here.

Issues related to the scarcity of \emph{car} window examples were alleviated by over-sampling of \emph{car} regions at training. In addition to the user-defined ROI, partial \emph{car} windows whose intersection with said ROI was greater than a $t_{ROI}$ threshold value were also considered (Fig.~\ref{fig:AllCarROIOversampling_Examples}). This approach had two advantages: i) it enabled training balanced classifiers with large number of examples, and ii) encouraged the classifier to be invariant to the position of the sampled windows in relation to the \emph{car} ROI. $t_{ROI}$ was set to 0.5 for square $512{\times}512$ windows (Fig.~\ref{fig:AllCarROIOversampling_Examples}.A) and to 0.65 for ${350\times}1050$ rectangular window, increasing the number of \emph{car} windows examples available at training by factors of ${\approx}140$ and ${\approx}50$, respectively (Fig.~\ref{fig:AllCarROIOversampling_Examples}.B).

\begin{figure}
\centering
\includegraphics[width=0.8\linewidth]{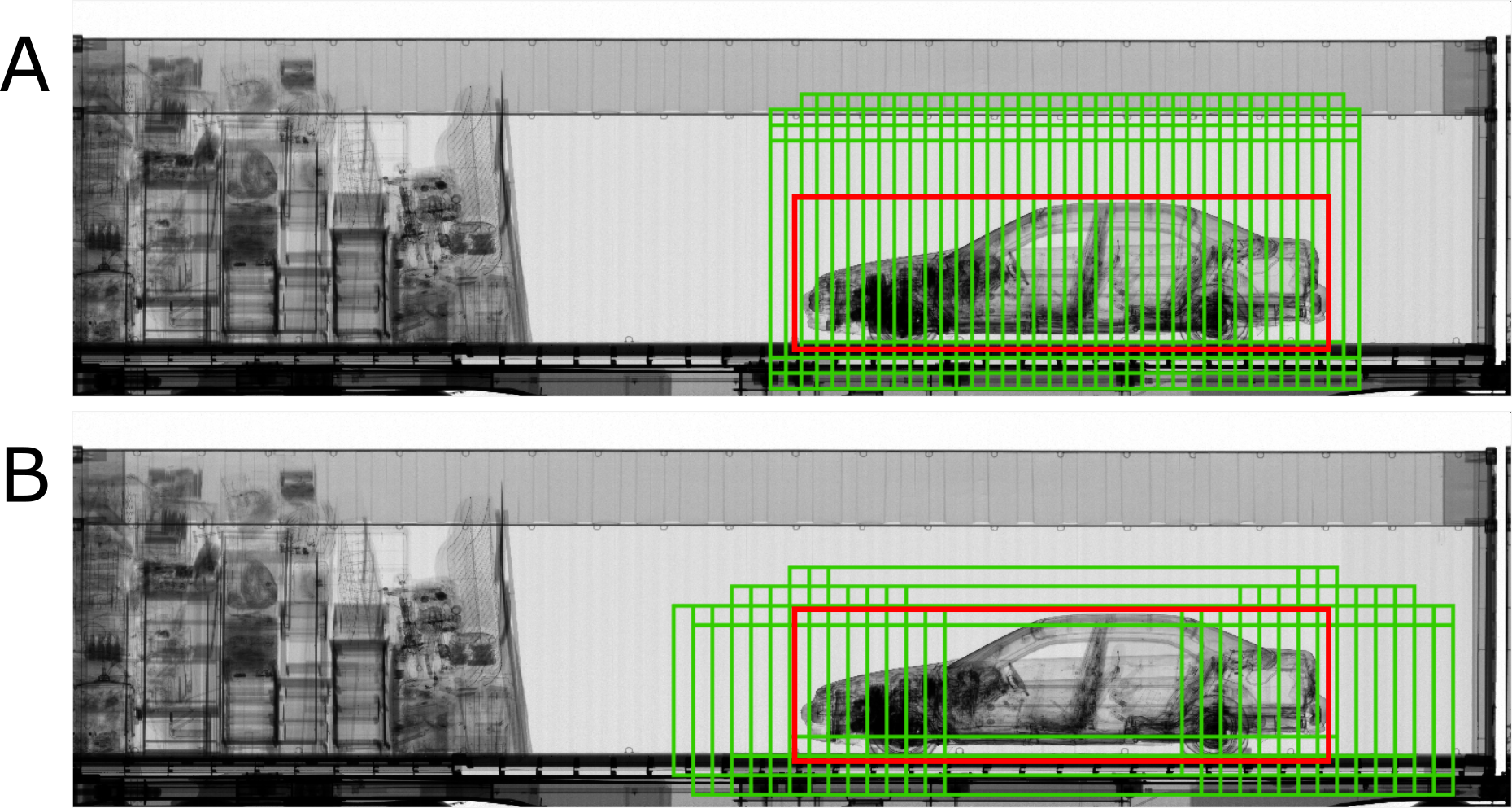}
\caption{Example of \textit{car} windows over-sampling. Windows in green are over-sampled and red windows indicate the user-annotated region of interest. Panels A and B show square window with $t_{ROI}=0.5$ and rectangular windows with $t_{ROI}=0.65$, respectively.}
\label{fig:AllCarROIOversampling_Examples}
\end{figure}

\subsection{Performance evaluation}
Performance was evaluated on the classification of entire images as \emph{car} or \emph{non-car} based on aggregated window scores. Two assumptions were made: (i) \textit{non-car} images (negative class) were generally associated with lower $p_{I}$ values (image score) than \textit{car} images (positive class); and (ii) achieving high detection rate on \textit{car} images was trivial but doing so while minimizing false alarms on \textit{non-car} (e.g. high sensitivity, high specificity classification) is challenging. \emph{Non-car} images were partitioned into disjoint training, validation, and test sets each comprising 10,000  SoC images. 

The performance evaluation scheme was identical across all combinations of features and classifiers. Leave-one-out cross-validation (LOOCV) was used for the determination of $p_{I}$ for \emph{car} images due to the low number of examples of the positive class in the dataset. A classifier was trained using windows sampled from 78 \textit{car} images and the \textit{non-car} training set before being used to infer $p_{I}$ for the left-out \textit{car} image. The $p_{I}$ for \emph{non-car} validation images was computed using a classifier trained on all 79 \emph{car} images and the same \emph{non-car} training images. All free parameters, including $t_{CAR}$, were then tuned before repeating the process, with fixed parameters, using the \emph{non-car} test images.

Combining the $p_{I}$ values obtained for the negative class (hold-out on validation or test set) and positive class (LOOCV), performance metrics such as the area under the ROC curve (AUC) and the H-measure could be computed. The latter was introduced by Hand and Anagnostopoulos~\cite{Hand2012} to suitably accommodate imbalanced datasets, such as the one considered here, while also addressing issues related to the underlying cost function of the AUC metric. Like the AUC, the H-measure can be computed without having to explicitly set a value for the threshold parameter (here $t_{CAR}$). A beta distribution with modes ($\pi_{2} + 1$, $\pi_{1} + 1$) is used as distribution of relative misclassification severities, where $\pi_{2}$ and $\pi_{1}$ are the relative frequencies of the positive and negative class, respectively. Details regarding the H-measure computation are given elsewhere~\cite{hand2009measuring} and implementations for most scientific computing packages are freely available\footnote{http://www.hmeasure.net/ - Last accessed 23.06.2016}. The false positive rate (FPR) was computed by thresholding the test set $p_{I}$ scores using the highest possible value for $t_{CAR}$ (tuned individually for each experiment based on validation images) that still resulted in 100\% \emph{car} image classification accuracy.

During performance evaluation, dictionary learning for PHOW features and mean image computation for CNNs were carried out solely based on training images (e.g. new dictionaries were learned and new mean images were computed for each iteration of LOOCV).

\subsection{Generation of synthetically obscured car examples}
Synthetically obscured \emph{car} images were generated by projecting \emph{non-car} objects onto SoC \emph{car} images. Due to the nature of the X-ray transmission image formation process, objects can be inserted into images by multiplication as previously described by Rogers \emph{et al.}~\cite{rogers2015detection}. The process started with a raw \emph{car} image. A first object was sampled from a database containing a total of 196 objects and placed at a random location in the container. The dimensions and density of the object were set to half and a third of that of a typical \emph{car}, respectively. The newly generated synthetic image was then classified and the image score $p_{I}$ computed. The mean relative attenuation of the \emph{car} ROI was computed as the difference between the synthetic image and the raw image, divided by the raw image. This process was repeated, adding more and more objects, until the car was completely obscured (mean relative attenuation equal to one). Five different realisations of this experiment were combined to generate a plot of the image score versus mean relative attenuation.
\section{Results}
\label{sec:results}
For each type of feature considered, the best car image classification results obtained across different combinations of pre-processing, window geometry and classifiers are presented in table~\ref{tbl:results}. It was found that an approach combining multi-scale computation (scale${=}\{1,2,4,8\}$) and encoding using 256-bin histograms (though diminishing returns were observed from 32-bin upwards) was optimal for intensity features. Log-transforming windows prior to analysis was found to be detrimental but using rectangular windows (based on prior knowledge about car geometry) significantly improved performance over square windows (H-measure of 0.95 and 0.86, respectively). However, intensity features performed the worst when compared to other types of features with a false alarm rate above 5\%; while the differences in intensity distribution between \emph{car} and \emph{non-car} windows might be a useful cue for classification, more advanced image descriptors such as PHOW and oBIFs were required to achieve satisfactory levels of performance. 

PHOW features outperformed intensity features when using raw images as input and log-transforming windows led to a further two-fold decrease in false alarm rate to approximately 1\%. Interestingly, oBIFs outperformed PHOW features even though the former do not rely on \emph{ad-hoc} dictionary learning or a pyramidal scheme. Instead, oBIFs are fixed geometric descriptors computed independently at multiple scales. oBIFs results showed a $\approx$3-fold improvement in false alarm rate to 0.35\% when compared to PHOW features. Using BIFs instead of oBIFs led to a marked degradation in performance, indicating that orientation quantisation was beneficial for classification. Log-transforming input windows also had a negative impact on classification using oBIFs, which was potentially caused by the lack of apparent texture and structure in these transformed images.

\begin{table}[h]
\caption{Performance for the detection of cars in X-ray cargo images. Only the best results for each type of features shown. +Log denotes that input images were log-transformed prior to features computation. R and S denote $1050{\times}350$ and $512{\times}512$ windows, respectively.}
\label{tbl:results}
\resizebox{\columnwidth}{!}{%
\begin{tabular}{ l c c | c c}
\hline 
Features & Windows & Classifier & H-measure & FPR [\%] \\ 
\hline
Intensity (4 scales) & R & RF & 0.900 & 5.20\\
PHOW (4 scales) + Log & S & RF & 0.977 & 1.05 \\
oBIFs (4 scales, 2$\epsilon$) & R & RF & 0.992 & 0.35 \\
CNN 11-layer + Log & S & SM & 0.990 & 0.47 \\ 
\textbf{CNN 18-layer (FC1) + Log} & \textbf{S} & \textbf{RF} & \textbf{0.995} & \textbf{0.22} \\ 
ImageNet VGG-VD-19 (FC2) + Log & S & SVM & 0.993 & 0.34 \\
\hline
\end{tabular} 
}
\end{table}

The best performance across all experiments, correct classification of all cars and a false positive rate of 0.22\% ($p_{I}$ = 0.990), was achieved using features extracted from the FC1 layer of a trained-from-scratch CNN when square input windows were log-transformed and classification was carried out using a random forest model. The 95\% confidence interval for the detection rate, which was estimated by supplementing the results with a single artificial failure case, was [0.96, 100].

The 18-layer trained-from-scratch CNN outperformed the shallower 11-layer network in all cases, indicating that the former generalised well to unseen data despite significantly increased complexity. The second-best result was obtained using a CNN pre-trained on the ImageNet dataset with no further fine-tuning, which suggests that features learned from natural images constitute a robust baseline for X-ray image classification. 

\begin{figure}[h]
\centering
\includegraphics[width=0.95\linewidth]{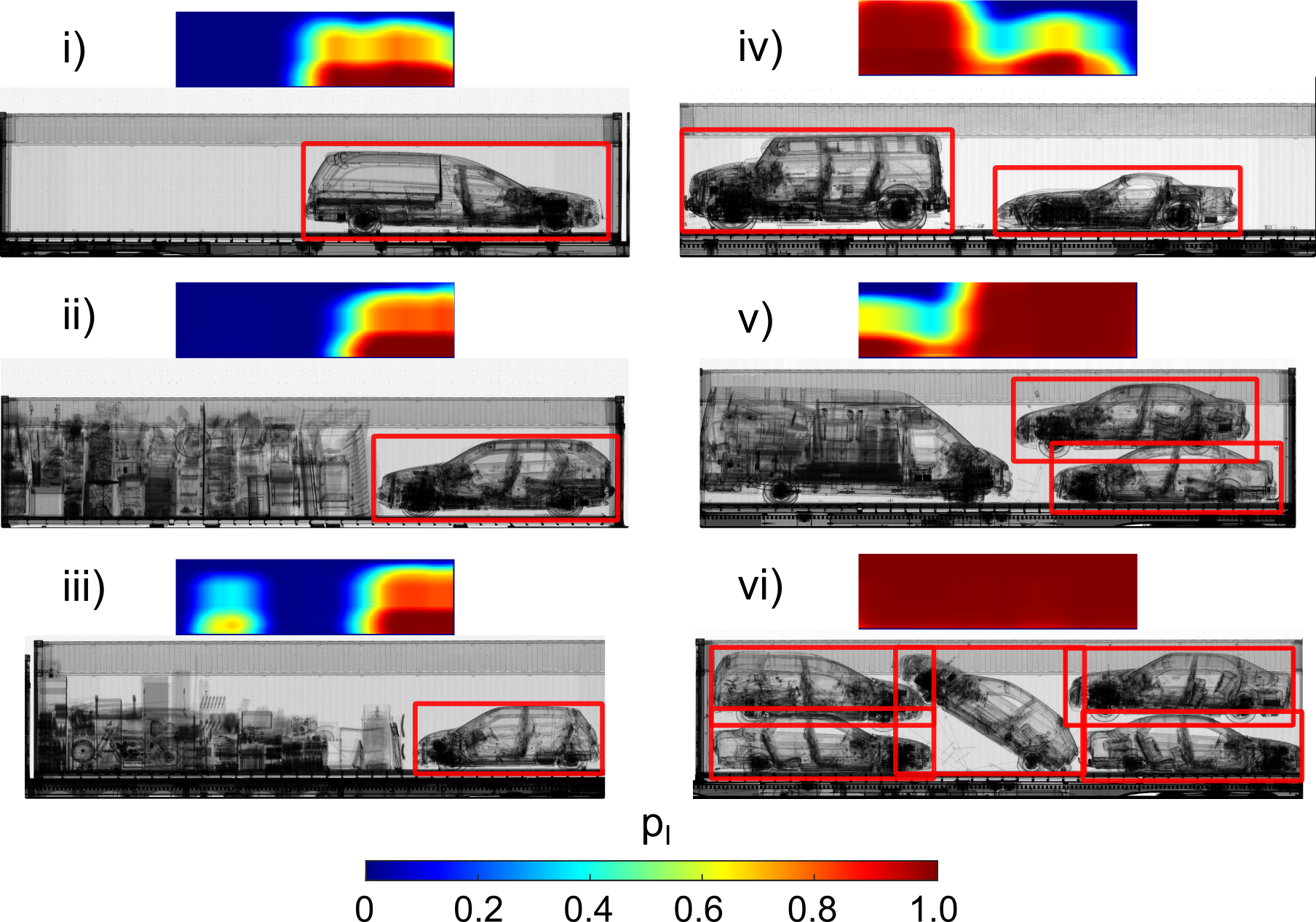}
\caption{Classification outcome for \textbf{non-obscured} \emph{car} images during leave-one-out cross-validation (previously unseen by classifier). For each example, raw X-ray transmission image (top, with additional red outlines indicating the location of cars) and the output of the classifier formatted as a heatmap (bottom) are shown.}
\label{fig:Cars_NotObscured}
\end{figure}

Figure~\ref{fig:Cars_NotObscured} shows representative examples of \emph{car} image classification by the CNN scheme where individual cars are not obscured by other goods. Various scenarios are shown: single cars without other goods (Fig.~\ref{fig:Cars_NotObscured}.i), multiple cars without other goods (Fig~\ref{fig:Cars_NotObscured}.iv, v, and vi), car with other goods (Fig.~\ref{fig:Cars_NotObscured}.ii and iii), cars with other vehicles (Fig.~\ref{fig:Cars_NotObscured}.v), and cars at an angle (Fig.~\ref{fig:Cars_NotObscured}.vi). In all cases, cars were also suitably localised by the heat map generated during classification regardless of the model (e.g. sedan, coupe, station wagon, SUV) and dimensions. Regions of images that contained other unrelated cargo usually gave very little to no signal (Fig.~\ref{fig:Cars_NotObscured}.ii), with the exception of cases where said cargo also included semantically-related objects, such as motorbikes (Fig.~\ref{fig:Cars_NotObscured}.iii) or vans (Fig.~\ref{fig:Cars_NotObscured}.v). The CNN scheme also performed well for complex X-ray imagery in which cars were partially and completely obscured by other cargo (Fig.~\ref{fig:Cars_Obscured}).

\begin{figure}[h]
\centering
\includegraphics[width=0.95\linewidth]{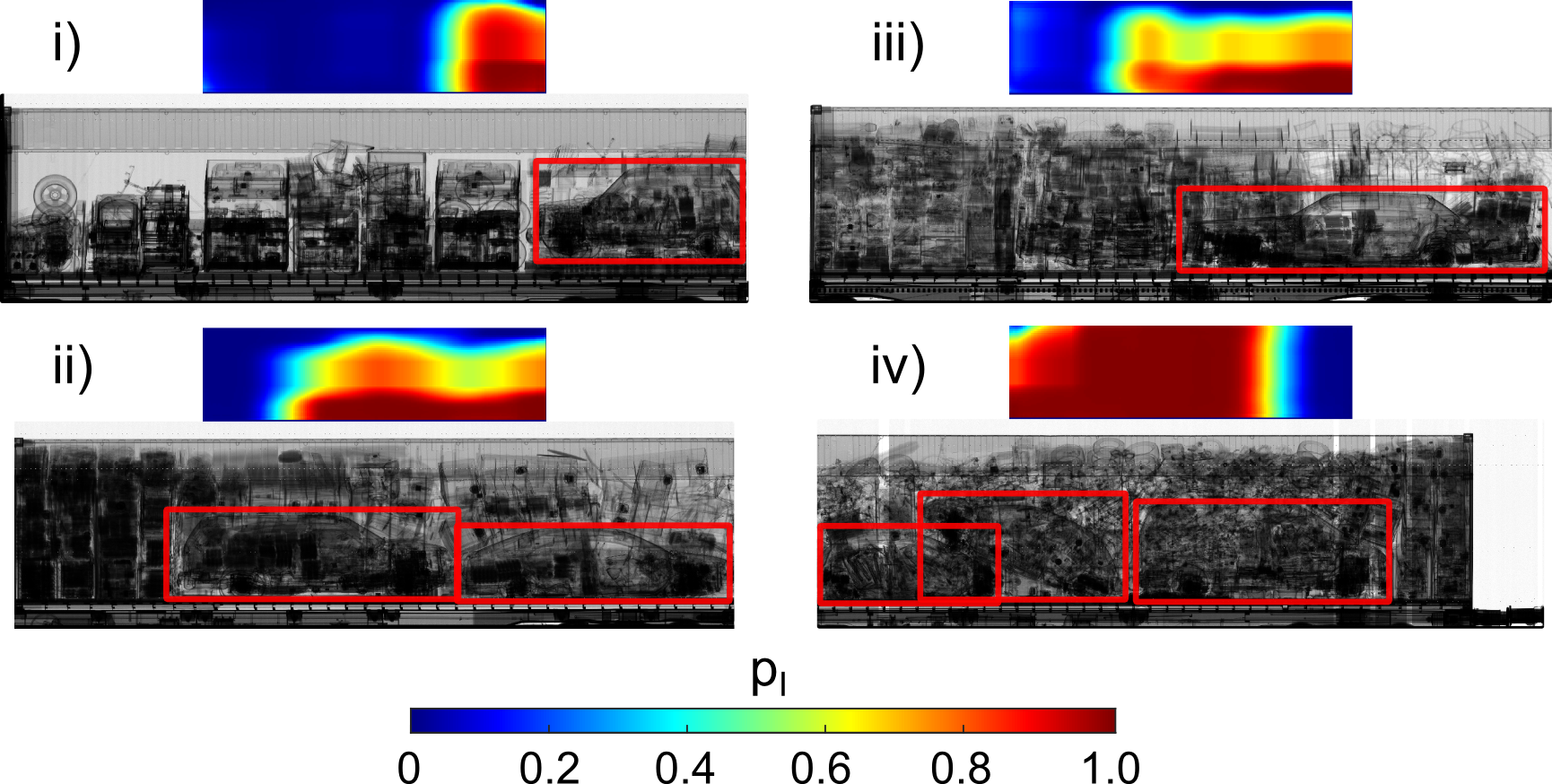}
\caption{Classification outcome for \textbf{obscured} \emph{car} images during leave-one-out cross-validation (previously unseen by classifier). For each example, raw X-ray transmission image (top, with additional red outlines indicating the location of cars) and the output of the classifier formatted as a heatmap (bottom) are shown.}
\label{fig:Cars_Obscured}
\end{figure}

The vast majority of \textit{non-car} images (97.82\% of the test set) had $p_{I}{\le}0.5$ and are thus correctly classified using a naive $t_{CAR}{=}0.5$ threshold (Fig.~\ref{fig:NonCarImages}). These images typically include empty containers (Fig.~\ref{fig:NonCarImages}.i a), containers fully filled with bulk materials (Fig.~\ref{fig:NonCarImages}.ii a), and containers with goods loaded onto pallets (Fig.~\ref{fig:NonCarImages}.iv a). The intermediate image category ($0.5{<}p_{I}{\le}0.95$, 1.6\% of the test set) was more challenging to classify due to the presence of uncommon high frequency structures. This includes industrial vehicles (Fig.~\ref{fig:NonCarImages}.i b), containers sparsingly loaded with bulk materials (Fig.~\ref{fig:NonCarImages}.ii b), and pallets containing  objects that present \emph{car}-like features. Images with $p_{I}{>}0.95$ represented 0.6\% of the test set and a majority of those contained objects that are visually and semantically similar to cars, including motorbikes (Fig.~\ref{fig:NonCarImages}.i c) and vans (Fig.~\ref{fig:NonCarImages}.ii b). Indeed, if only considering false alarms that are not related to vehicles, the FPR for the CNN scheme decreases from 0.22\% to just 0.08\%.

Interestingly, the response for a given type of objects differed vastly depending on factors such as orientation, spatial arrangement, and fraction of space left empty in a container. For example, bulk materials were usually associated with very low image scores when uniformly loaded throughout the entire container but result in scores that tended to increase as more container background was visible (see Fig.~\ref{fig:NonCarImages} ii a-c). Similarly, images of tires, which are semantically related to cars, were given low scores when their arrangements inside the container was such that a majority only had their profile showing (Fig.~\ref{fig:NonCarImages}.iii a). However, image score increased markedly as packing order decreased and tires adopted multiple orientations, including those typically seen in \emph{car} images (Fig.~\ref{fig:NonCarImages}.iii b and c).

\begin{figure}[h]
\centering
\includegraphics[width=0.95\linewidth]{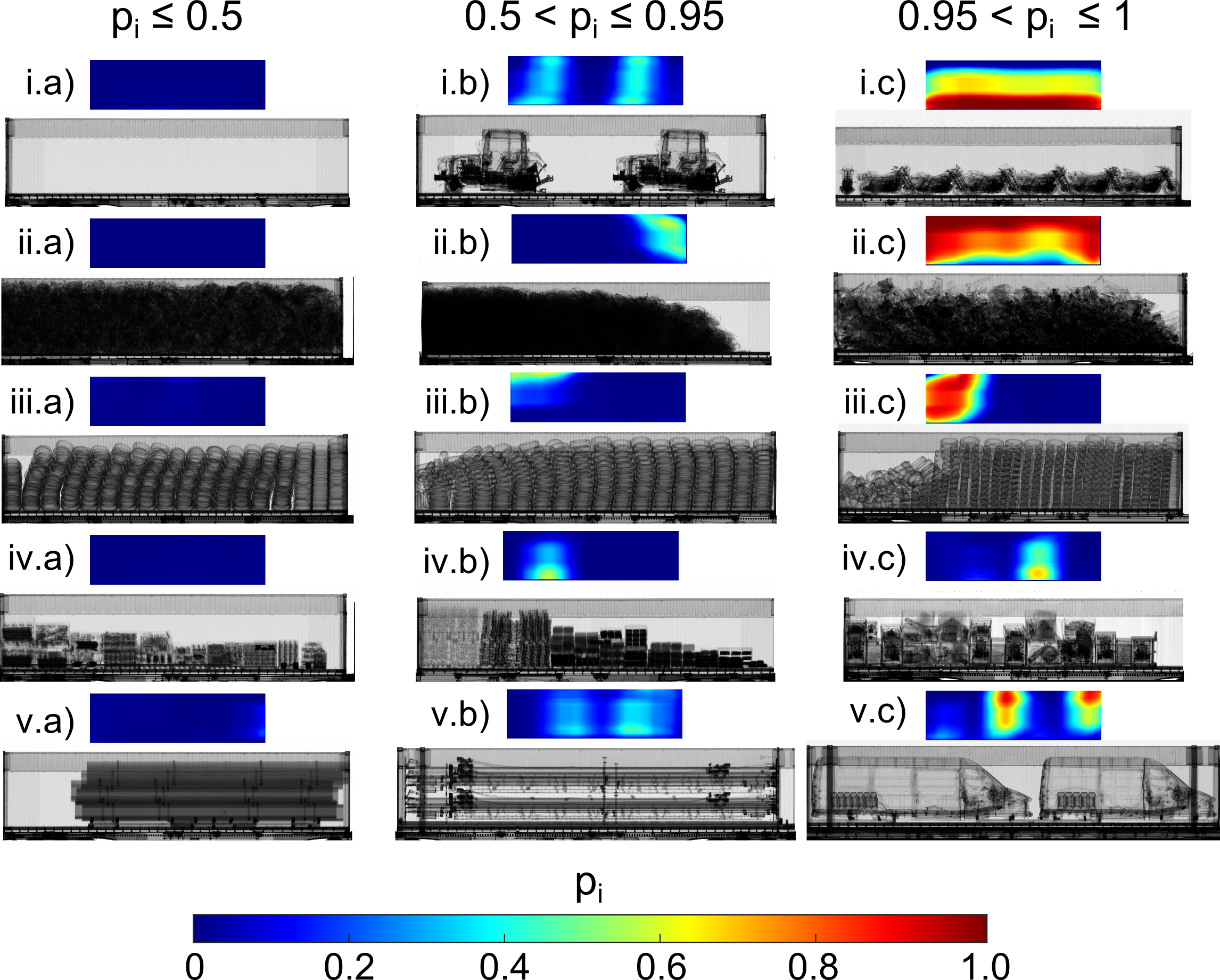}
\caption{Classification outcome for SoC \textit{non-car} images from the test set (previously unseen by the classifier). For each example, the X-ray transmission image (bottom) and the localisation heatmap (top) are shown.}
\label{fig:NonCarImages}
\end{figure}

The robustness of the proposed classification scheme to obscuration of cars was evaluated by generating synthetic adversarial images where other goods were projected into \emph{car} images (Fig.~\ref{fig:obfuscationAndBands}). Up to a mean relative attenuation of 0.8, which corresponds to a visually very busy scene, the CNN scheme consistently classified the synthetic images as \emph{car} (i.e. $p_{I}{\ge}0.990$). This indicates good resilience to concealment strategies commonly used by criminals as shielding methods to provide this degree of attenuation would be difficult to devise in practice. The spatial arrangement of the obscuring objects also played a role as shown by distinct synthetic images with the same relative attenuation value resulting in different classification outcomes.

\begin{figure}[h]
\centering
\includegraphics[width=0.55\linewidth]{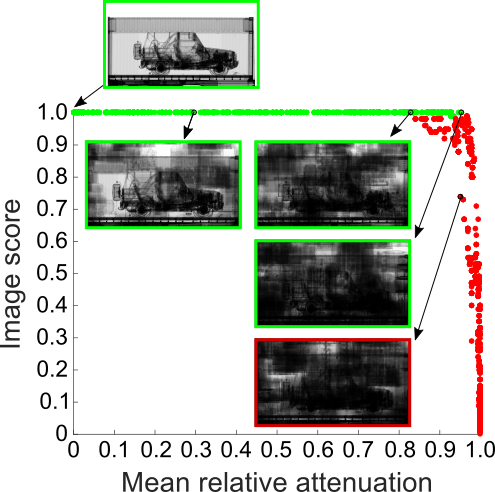}
\caption{Evaluation of the robustness of the classification scheme to obscuration by other cargo. The green and red points indicate scores above and below the optimal threshold $p_{I}$ = 0.990, respectively.}
\label{fig:obfuscationAndBands}
\end{figure}

\section{Conclusion}
\label{sec:conclusion}
We described a scheme for the detection of cars in X-ray cargo imagery whereby densely sampled windows were classified using Convolutional Neural Networks (CNNs). The proposed approach significantly outperformed other methods such as classification of windows based on intensity features, Pyramids Histograms of Visual Words (PHOW), and oriented Basic Image Features (oBIFs). \textit{Car} window oversampling alleviated issues associated with the low number of \textit{car} examples and enabled the training of networks from scratch instead of solely relying on pre-trained networks as done previously for baggage imagery~\cite{akccaytransfer}.

All \emph{car} images in the stream-of-commerce dataset were correctly classified as such, including cases where cars were partially or totally obscured by other goods. In an experiment based on synthetic images, the CNN scheme demonstrated a high degree of robustness to obscuration by accurately classifying images that could be deemed challenging for Human observers. This resilience might be made possible by the use of log-transformed input images; due to the nature of the X-ray image formation process, \emph{car}-like structures are preserved in those transformed images and can still act as classification cues that CNNs excel at picking up, even for moderate to high relative attenuation values.

When trained from scratch, CNNs can thus suitably accommodate properties of X-ray imagery, such as translucency and multiplicative occlusion, that do not typically occur in natural images. Moreover, with fewer than 1-in-450 false alarms, it was shown that perfect sensitivity did not come at the cost of unsuitably low specificity. A large fraction of those false positives were comprised of images containing objects semantically related to cars.

On average using a MATLAB implementation running on a E5-1620 Intel Xeon 3.4 Ghz CPU, a Titan X GPU and 32 GB of RAM, image classification including pre-processing, sampling, CNN features computation, and classification using Random Forest took 2.6 seconds. It is likely that processing could be further improved by taking advantage of lower-level coding, smart batch processing of sampled windows, and multiple GPU setups.

Several limitations and areas for future work were identified. The images used in the experiments were all acquired using a single X-ray machine. Further investigations will be required to determine how the processing parameters and classifiers would generalise across different models, operating conditions (e.g. photon energy), and imaging protocols. In particular, differences in image geometry (e.g. perspective, warping) might require instrument-specific calibrations. It would also be beneficial to evaluate performance on a larger dataset of \emph{car} images. However, while classification performance does increase with the number of \emph{car} images used at training, it was found that said performance started to plateau and adding more training examples from 50 images onwards yielded diminishing returns.

The reported performance suggests that this approach could be deployed in the field to assist operators in the detection of fraud and crime related to the undeclared transport of cars in cargo containers. Due to its generic nature, this deep learning scheme could likely be used to detect many classes of objects in complex X-ray imagery, even when only a modest dataset of examples is available.

\section{Acknowledgements}
This work was funded by Rapiscan Systems Ltd. and through the EPSRC Grant no. EP/G037264/1 as part of UCL’s Security Science Doctoral Training Centre.

\bibliography{NJ_Cars_ArxivVersion}

\end{document}